\pdfoutput=1

\documentclass[11pt]{article}

\usepackage{ACL2023} 

\usepackage{times}
\usepackage{latexsym}
\usepackage{float}
\usepackage[T1]{fontenc}

\usepackage[utf8]{inputenc}

\usepackage{microtype}

\usepackage{inconsolata}
\usepackage{graphicx}
\usepackage{comment}
\usepackage{booktabs}
\usepackage{colortbl}
\usepackage{tabulary}
\definecolor{greenish}{HTML}{def5e5}
\definecolor{orangy}{HTML}{F0FFFF}

\title{On the Cross-lingual Sharing of Cultural Values: Stability and Training Data Influence and something more interesting }
\title{
On the Cross-lingual Stability of Cultural Values Disentangling the Cross-lingual Sharing of Cultural Values}
\title{Different training setting: few-shot, in-context, how many in-context examples would we need in the target language until the label flips
}

\title{TRAKing changes in cross-cultural values in multilingual LMs during fine-tuning}
\title{Which cultural values do languages transmit to multilingual LMs?\\ TRAKing cross-cultural values during LM fine-tuning}
\title{Which cultural values do languages transmit to multilingual LMs?\\ TRAKing the emergence of cultural values back to their source}

\title{How are cultural values transmitted to multilingual LMs?\\ TRAKing changes in cultural values back to their data source}

\title{How multicultural are multilingual LMs?\\ Tracing cultural changes back to their data source}

\title{How do cross-cultural values interact within multilingual LMs?\\ Tracing cultural changes back to their data source}

 \title{How do cross-cultural values evolve in multilingual LMs?\\ Tracing cultural changes back to their data source}

 \title{How does cross-language influence affect cross-cultural value shifts\\ in multilingual LMs?}

 \title{How different languages influence the encoding of cultural values in multilingual language models?}

 \title{Tracing Cross-cultural Value Shifts in Multilingual LMs}
 
 \title{Tracing Cross-language Influence on \\ Cultural Value Shifts in Multilingual LMs}

\title{The Echoes of Multilinguality: \\Tracing Cultural Value Shifts during LM Fine-tuning}
 
\author{Rochelle Choenni \\ University of Amsterdam \\ \texttt{r.m.v.k.choenni@uva.nl} \And Anne Lauscher \\ University of Hamburg \\ \texttt{anne.lauscher@}\\\texttt{uni-hamburg.de} \And Ekaterina  Shutova \\ University of Amsterdam \\ \texttt{e.shutova@uva.nl}}

\begin{document}
\maketitle
\begin{abstract}
Texts written in different languages reflect different culturally-dependent beliefs of their writers. Thus, we expect multilingual LMs (MLMs), that are jointly trained on a concatenation of text in multiple languages, to encode different cultural values for each language. Yet, as the `multilinguality' of these LMs is driven by cross-lingual sharing, we also have reason to belief that cultural values bleed over from one language into another. This limits the use of MLMs in practice, as apart from being proficient in generating text in multiple languages, creating language technology that can serve a community also requires the output of LMs to be sensitive to their biases \citep{naous2023having}. Yet, little is known about how cultural values emerge and evolve in MLMs \citep{hershcovich-etal-2022-challenges}.
We are the first to study how languages can exert influence on the cultural values encoded for different test languages, by studying how such values are revised during fine-tuning.
Focusing on the fine-tuning stage allows us to study the interplay between value shifts when exposed to new linguistic experience from different data sources and languages. Lastly, we use a training data attribution method to find patterns in the fine-tuning examples, and the languages that they come from, that tend to instigate value shifts.     
\end{abstract}

\section{Introduction}
Training LMs on large text corpora has been shown to induce various types of (social) biases in multilingual LMs (MLMs) 
that affect which human values the model picks up on~\citep{choenni2021stepmothers, hammerl2022speaking}.  
However, human values vary per culture, which means that the cultural values that are reflected through their language (either explicitly or implicitly) will also differ. As MLMs are trained on the concatenation of text from a wide variety of languages spoken in the world, we can expect different, and perhaps opposing, cultural values to be encoded in them simultaneously. 
This necessitates MLMs to become inherently multicultural as well in order to appropriately serve culturally diverse communities~\citep{naous2023having, talat2022you}. In fact, it has already been shown that MLMs encode a distinct set of cultural values for different languages. 
However, those values do not tend to align with those collected from real human surveys conducted in the countries where the majority population speak the respective languages \citep{arora2023probing, kovavc2023large}. As such, the multilingual NLP community is now faced with the new pressing challenge of better culturally aligning MLMs to human values~\citep{yao2023instructions}. Thus, we aim to study how cultural values emerge and evolve in MLMs to better understand and aid cross-cultural value alignment in the future.

\begin{figure}[!t]
    \centering
    \includegraphics[width=\linewidth]{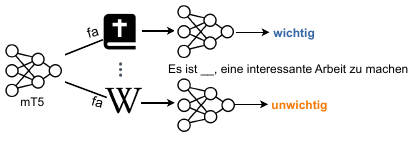}
    \vspace{-0.6cm}
    \caption{An example of our evaluation setup. We analyze the effect that fine-tuning on different data sources in a fine-tuning language $A$ (Farsi) has on the cultural values that are encoded for a test language $B$ (German). Translation of the example: \emph{`It is \_ to do interesting work.'} (options: \emph{`important'} or \emph{`unimportant'}).} 
    \label{fig:method}
\end{figure}

 In particular, we hypothesise that training on multilingual data leads to an interaction of language-specific cultural values within the models, possibly steering a language's cultural bias into a direction that is unfaithful to the majority of that language's speakers. This raises an interesting set of questions on how languages exert influence on the encoding of cultural values. Focusing on the fine-tuning stage, we study how cultural values are revised during training. For instance, given a set of fine-tuning languages, test languages, and data sources, when we fine-tune on a data source in a language $A$, and test in a language $B$, do we inadvertently induce the cultural values from $A$ into $B$? And would the same effect be visible across all test languages or are the values encoded for some languages more prone to change? Moreover, how much impact does the bias of the data source itself have versus the language used for fine-tuning? 
 And can different data sources systematically enforce different alignments to human values? 

To better understand this cross-cultural interaction, we study the following questions: 
(\textbf{Q1}) How do the fine-tuning language and data source affect the way in which cultural information is encoded and revised during fine-tuning?
(\textbf{Q2}) How do cultural value shifts 
change the alignment to human survey data? 
(\textbf{Q3}) Can we find patterns in the training examples that systematically influence how cultural values are revised? 
To this end, we conduct a set of controlled fine-tuning experiments using multi-parallel texts from data sources with neutral bias (Wikipedia), religious bias (Bible and Quran) and political bias (news articles) across 4 fine-tuning languages and 13 test languages. We follow \citet{arora2023probing} in using 200$+$ WVS survey questions to probe for cross-cultural values in pretrained and fine-tuned MLMs. Importantly, using survey questions as probes allows us to test the alignment between model predictions and human data. 
Finally, we use a training data attribution (TDA) to trace value shifts back to the data source. 

 We find that, while fine-tuning language and domain source play a minor role in the revision of cultural information compared to the amount of fine-tuning data, fine-tuning languages can lead the cultural perspective of test languages into different directions. Importantly, this can positively affect the models' alignment to human values. Yet, overall, results vary considerably across test languages. Moreover, our TDA analysis provides interesting insights about the systematicity with which the model tends to rely on parallel data to instigate the same value shifts across languages. Our results underpin the complexity of cross-language and cross-cultural interaction within MLMs, and suggest that the semantic content of fine-tuning data might not be the main reason for value shifts. 

\section{Related work}

\subsection{Cross-cultural NLP}

The fact that LMs are becoming increasingly multilingual, has given rise to a new subfield in NLP that is concerned with questions such as to what extent these models are multicultural ~\citep{liu2023multilingual, havaldar2023multilingual, hershcovich-etal-2022-challenges}, to what extent the cultural values that they encode align with those from human populations~\citep{naous2023having, arora2023probing, cao2023assessing}, and whether we can automatically improve such an alignment to better serve culturally diverse communities~\citep{kovavc2023large}. For instance, \citet{naous2023having} show that MLMs tend to exhibit western-centric biases, even when being prompted in Arabic and contextualized by an Arab cultural
setting, resulting in culturally insensitive output such as suggesting to go for a beer after Islamic prayer. Similarly, previous works show that LMs fail to understand proverbs and sayings from different languages~\citep{liu2023multilingual}, and do not capture the nuances in
meaning and usage patterns of emotion words that exist differently across cultures~\citep{havaldar2023multilingual}. These findings suggest that there is still an important gap to fill when it comes to creating multilingual language technology that is also multicultural~\citep{talat2022you}. We aim to contribute to our understanding of how cultural values manifest across languages. 

\subsection{Probing for bias}

Cloze-style testing is a technique that stems from psycholinguistics \citep{taylor1953cloze}, and has been popularized as a tool to study different types of knowledge and biases encoded by LMs. The idea is that we prompt LMs with a carefully curated set of probing sentences that are meant to elicit responses that expose the biases encoded within the LM \citep{may2019measuring, stanczak2023quantifying, nangia2020crows}. While many different types of biases have been studied in the multilingual setting~\citep{hammerl2022speaking, touileb2022occupational, kaneko2022gender}, \citet{arora2023probing} are the first to probe for cross-cultural values in pretrained MLMs. We use their probing questions in a similar set-up, but take a step further by studying how different fine-tuning languages can exert influence on cultural values encoded for a different set of test languages.

\subsection{Training data attribution}
Training data attribution (TDA) methods are developed to identify a set of training examples that were most influential in making a particular test prediction. The influence of a training example $z_{train}$ on test example $z_{test}$ can typically be formalized as the change in loss that would have been incurred for $z_{test}$, if sample $z_{train}$ was not seen during training \citep{koh2017understanding}. Thus, we can use the resulting influence scores as a measure of how important $z_{train}$ is for making a prediction for $z_{test}$. TDA methods have successfully been used on classification tasks in NLP~\citep{han2022orca}
, e.g.,\ to detect outlier data \citep{han2020explaining, lam2022analyzing} 
 or to correct model predictions \citep{meng2020pair, guo2021fastif}. Recently, they have been applied to study cross-lingual sharing in MLMs \citep{choenni2023languages, choenni2023examining}. Yet, extending the use of TDA methods beyond classification tasks 
has proved difficult. \citet{akyurek2022tracing} first used 
TDA methods~\citep{rajani2019explain, pruthi2020estimating} on masked language modelling for fact tracing -- the task of attributing an LM’s factual assertions back to training examples. Yet, the results were shown to be unreliable. More recently, however, \citet{park2023trak} proposed TRAK, which was shown to be successful in \emph{behaviour tracing} on mT5. We adopt their approach to trace mT5 predictions for cloze-style questions eliciting cultural values back to the fine-tuning data.

\section{Methodology}

\subsection{World Values Survey (WVS)}
We probe for cultural values using 
cloze-style testing templates derived from the questions proposed in the World Values Survey (WVS)~\citep{haerpfer2022} by \citet{arora2023probing}. Thus, more precisely, we study descriptive ethics~\citep{vida2023values}.
The WVS collects data on cultural values in different countries in waves, 
and our questions come from Wave 7 which ran from 2017 to 2020 and targets 57 countries~\footnote{\href{https://www.worldvaluessurvey.org/WVSDocumentationWV7.jsp}{https://www.worldvaluessurvey.org}}. Survey results are published per question, organised in 13 categories: (1) Social
Values, Attitudes and Stereotypes, (2) Happiness and Well-being, (3) Social Capital, Trust and Organisational Membership, (4) Economic Values, (5) Corruption, (6) Migration, (7) Security, (8) Post-materialist Index, (9) Science and Technology, (10) Religious Values, (11) Ethical Values and Norms, (12) Political Interest and Political Participation, (13) Political Culture and Regimes.
Categories (4) and (8) are excluded as their questions could not be converted into probes. We use 237 probes in total.

\subsection{Multilingual probes}\label{sec:mprobes}
We use the English probes that were professionally translated into the following 13 languages: Romanian, Greek, Urdu, Farsi, Tagalog, Indonesian, German, Malay, Bengali, Serbian, Turkish, Vietnamese and Korean, see Figure~\ref{fig:method} for an example. Note that these languages were carefully selected by \citet{arora2023probing} based on the following three criteria: (1) the languages can be mapped to one country covered by the WVS survey, (2) the languages are the official languages of the countries that they are mapped to, (3) the distribution of
the language's speakers can be primarily localized to a country or relatively small geographical region, and (4) all selected languages have at least 10K articles on Wikipedia such that the LMs have seen a sufficient amount of pretraining data.

\subsection{Models}
\citet{arora2023probing} report that cultural information is inconsistent across different pretrained LMs. Given recent trends on scaling LMs to tens of billions of parameters \citep{workshop2022bloom}, we study how model size affects cultural information instead. 
We probe mT5~\citep{xue2021mT5} small, base and large that contain 0.3B, 0.58B and 1.2B parameters. 

\subsection{Probing method}
To probe for cultural values, we query the mT5 models with the
cloze-style question probes from Section~\ref{sec:mprobes} using a conditional language generation head. More concretely, for each probing template, we replace the \texttt{[MASK]} token in the original probes with extra ID tokens for mT5, and 
 apply softmax over the logits of all tokens in the vocabulary $V$. We then take the log probability for the two candidate answers of the question, and take the option with the highest log probability as the final answer.

\subsection{Quantifying shifts in cultural profiles}\label{sec:cultural_profiles}
To compare overall cultural bias across languages and model sizes, we build `cultural profiles' based on their predictions for all WVS questions. Per question, we take the log probabilties of the respective answers and apply softmax to them to obtain the probabilities for selecting the first answer option for all $N$ questions. We then compile them into a $N$-dimensional vector, which represents the cultural profile of a given language. 
Similarly, we obtain ground truth profiles for the corresponding countries using the results from the WVS survey. The results are reported as the percentage of interviewees that selected each class. Yet, in contrast to our probes, the survey proposes multiple classes (e.g.\ \textit{`very important'}, \textit{`important'}, \emph{`not very important'}, \emph{`not important'}). We add the probability from the middle classes to the closest class on either end of the spectrum e.g., very important/important becomes one class. We then test how similar cultural profiles within pretrained models are to the ground truth, and in which direction they change after fine-tuning, by computing the change in correlation.

\section{Experimental setup}

\begin{table}[t!]
    \centering   
    \setlength{\tabcolsep}{4pt}
    \scalebox{0.7}{
    \begin{tabular}{r | c c c c c c c c c c c c c |c}
    \toprule
                 & bn & de & el & fa & id & ko &ms & ro& sr & tl & tr & ur & vi&  avg. \\
            \midrule 
            S vs. B & 78&89 & 84& 89 & 77& 82& 84& 94&72 & 70& 79& 87& 83& 82 \\
            S vs. L & 78&87 & 83&  89&  69& 86& 81& 96&63 & 69& 75& 83& 83&  80\\
            B vs. L & 88& 86&  81& 87& 79&  83& 84& 94& 72& 77& 81& 85 &87 & 83\\
\bottomrule
\end{tabular}}

    \caption{Percentage of agreement between model predictions from the pretrained mT5 \textbf{S}mall, \textbf{B}ase and \textbf{L}arge models fine-tuned on PBC (10K).}
    \label{tab:sizes}
\end{table}

\subsection{Data sources}

We use three data sources with multi-parallel data for fine-tuning: Flores-101~\citep{goyal2022flores}, the Parallel Bible Corpus (PBC)~\citep{christodouloupoulos2015massively}, and the Tanzil dataset~\citep{tiedemann2012parallel} that contain human translated sentences from Wikipedia articles, Bible texts and Quran texts respectively. While Flores-101 is more likely to be used in practice, PBC and Tanzil are an interesting testbed 
as due to their didactic nature, we expect cultural values to be affected more heavily. 
We select 4 languages for fine-tuning: Farsi, Korean, Hindi and Russian. These languages all rely on a different writing script, and are commonly spoken by culturally diverse populations. Also, Farsi and Korean are included in our test languages. The PBC dataset already contains multi-parallel sentences, and for the Tanzil we were able to extract them automatically using the English sentences in the translation pairs.   
Finally, following \citet{choenni2021stepmothers}, we also fine-tune on articles from different news sources across the political spectrum from left to right-wing ideologies. We use English news articles collected between 2013 and early 2020 from New Yorker (\textit{left}), Reuters (\textit{center}) and FOX news (\textit{right}) from the \textit{All-The-News} dataset. While the focus of this study is on language influence, we use this as an additional test to disentangle the effect of language bias from domain bias.

\subsection{Training details}
From each data source we use either 2K or 10K consecutive sentences for fine-tuning on the MLM `span corruption' objective that was used for pretraining, see Appendix~\ref{app:FT} for training details. We use two fine-tuning strategies (FT): (1) monolingual FT, where we train our models on each language separately, and (2) multilingual FT, where we jointly train on all fine-tuning languages together. 
For (2), we use 2.5K multi-parallel sentences for each language and shuffle them before training. We compare multilingual and monolingual models where 10K sentences are seen in total.

\section{Probing results}
As a baseline to our fine-tuning experiments, we first study the cultural profiles encoded in the pretrained models. In Section~\ref{res:shifts}, we then analyze how cultural information in the model changes as a result of cross-language and domain influence.

\subsection{Cultural information in pretrained LMs}\label{sec:prob_results}
 As explained in Section~\ref{sec:cultural_profiles}, we build cultural profiles for each country and compute Spearman correlation between the ground truth and pretrained model profiles. In line with previous results \citep{arora2023probing}, we confirm that all pretrained LMs correlate poorly with human values. Yet, in Table~\ref{tab:sizes}, we find that the models of varying sizes on average agree on 80$\%$ of the survey questions (pairwise). 
 In addition, we find that variations in consistency mostly depend on the test language. For instance, in Romanian the models agree on $\geq94\%$ of questions, but for Serbian this is $\leq72\%$ instead. Similarly, averaged across test languages, the models agree more on specific WVS categories e.g., predictions are more consistent for questions pertaining to happiness, security and political culture ($\geq85\%$) and less consistent when it comes to ethical values, political interest and corruption ($\leq76\%$), see Appendix~\ref{app:pretrained}. As all models exhibit similar behavior we focus analysis on mT5-small. 

\subsection{Cultural value shifts}\label{res:shifts}
In Section~\ref{sec:howmuch}, we study how the interplay between fine-tuning language, test language and data source will affect the \emph{amount} of value shifts. In Section~\ref{sec:whatdir}, we instead test how these factors more generally affect the cultural profiles across test languages by studying in which direction the models' bias changes. Finally, in Section~\ref{sec:monomulti}, we test how much cross-lingual sharing during multilingual fine-tuning will further impact these results.

\subsubsection{How big is the role of FT language and domain source on cultural value shifts?}\label{sec:howmuch}

\begin{figure}[!t]
    \centering

    \includegraphics[width=0.49\linewidth]{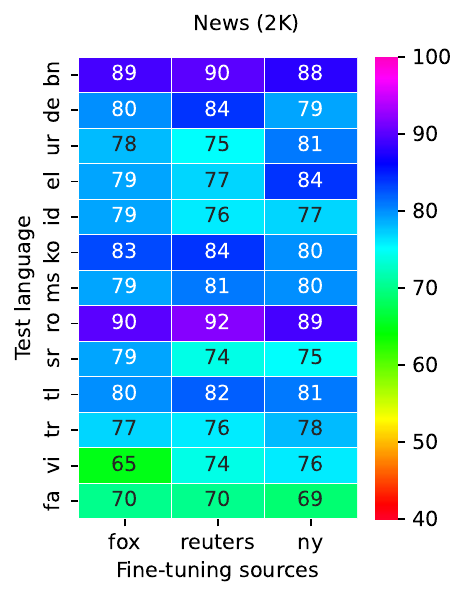}
        \includegraphics[width=0.49\linewidth]{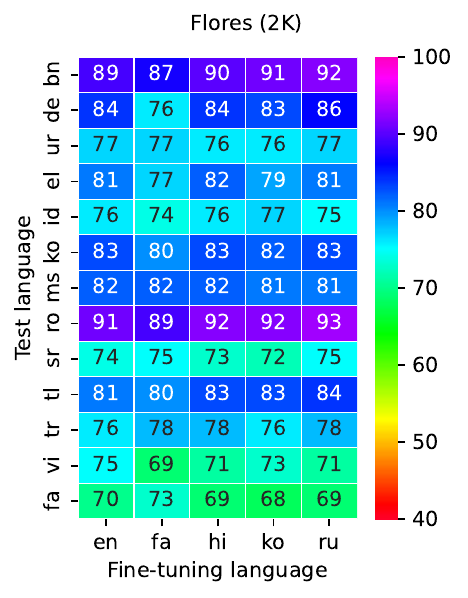}
        \includegraphics[width=0.49\linewidth]{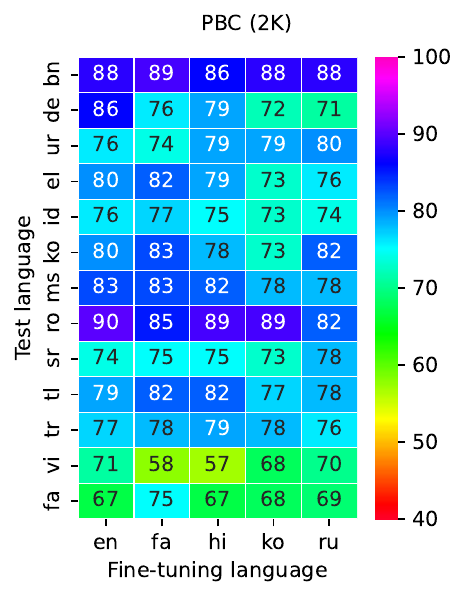}
        \includegraphics[width=0.49\linewidth]{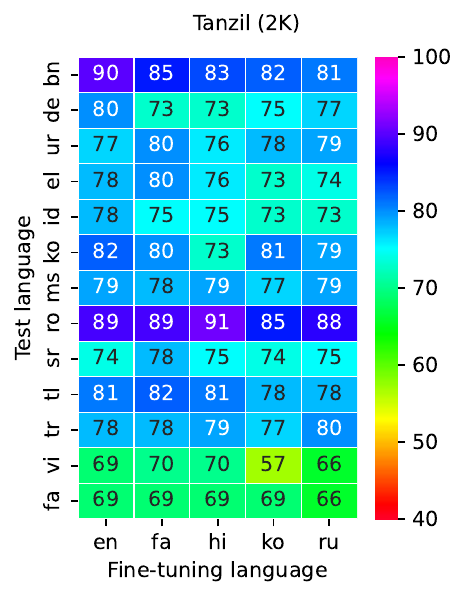}
         \includegraphics[width=0.49\linewidth]{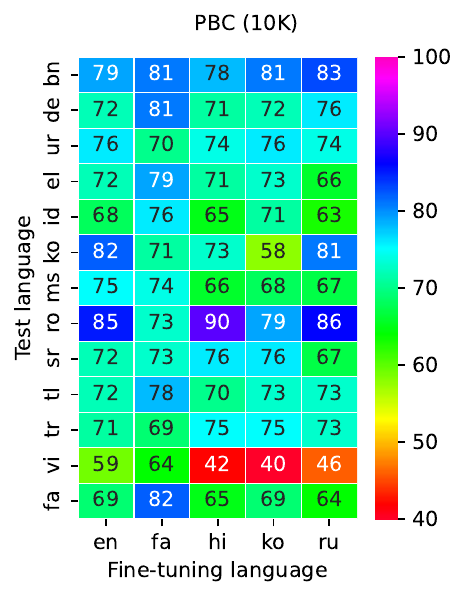}
        \includegraphics[width=0.49\linewidth]{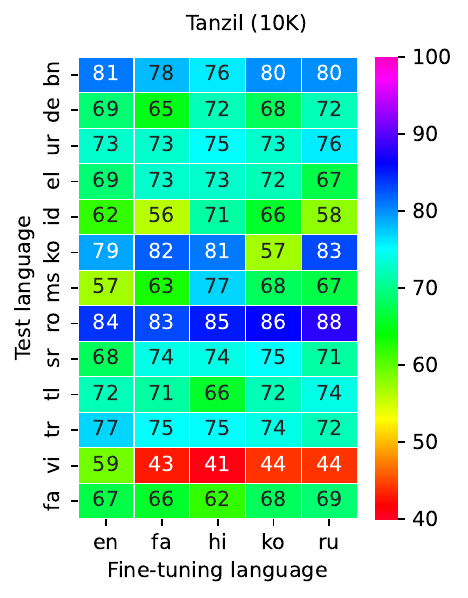}

    \caption{The percentage of predictions that remain unchanged after fine-tuning mT5-small.}
    \label{fig:flip_langs}
\end{figure}

 In Figure~\ref{fig:flip_langs}, we report the percentage of predictions that remain unchanged after fine-tuning on 2K sentences from news articles, Flores-101, PBC and Tanzil. We find that the amount of changes for Flores-101 are within the same ranges as for the news sources (7-35$\%$ shifts). In particular, we find that articles from Reuters (neutral bias) tend to result in the most value shifts. While this is somewhat surprising, it is line with our findings for Flores-101 that show shifts to similar extents. 
 However, as results across these data sources are less distinct in general, we focus most of our further analysis on PBC and Tanzil instead. 
 As expected, we find that PBC and Tanzil have a slightly larger impact on the cultural values encoded (9-43$\%$ shifts) than news articles and Flores-101 data.
 In particular, for PBC, fine-tuning in Korean and Russian have a bigger effect across test languages (e.g. for Greek and German). Similarly, when using Tanzil, next to Korean and Russian, Hindi has a larger effect as well. Yet, similar to our pretrained LMs, we find that the effect of fine-tuning language and source varies across test languages. For instance, for Farsi, regardless of the fine-tuning domain or source, many more values change than for the other languages.  
This shows that the effect of domain and language bias on the amount of value shifts is heavily dependent on the language that we study shifts for, making it difficult to draw general conclusions on which one has the largest impact overall. We, however, speculate that cultural information is separately encoded for each language in the model, and that the confidence with which these values are encoded varies depending on the test language.\footnote{If cultural values were jointly encoded across languages, we would expect cultural profiles to behave in a more homogeneous way given the same fine-tuning set up.}
Thus, based on the starting point, all fine-tuning setups will be able to affect the test languages to similar extents.


\begin{figure}
    \centering
    \scalebox{0.75}{
    \includegraphics[width=\linewidth]{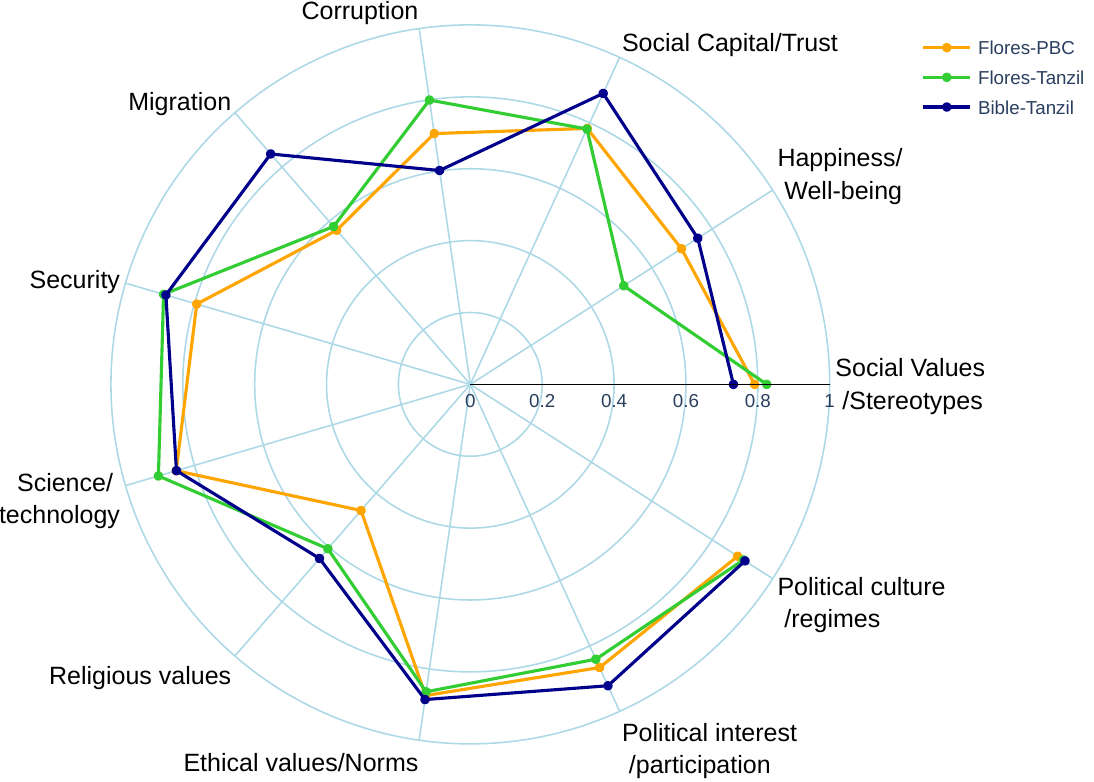}}
    \caption{Pearson correlation between the average percentage of unchanged values across fine-tuning languages for each data source pair per WVS category.}
    \label{fig:radar_cats}
\end{figure}

 \paragraph{Are certain cultural values more prone to shift?}

When studying the consistency with which values shift, we find that for each test language the values for the same questions tend to be affected, regardless of the fine-tuning language. Specifically, on average only 14$\%$ of the value shifts are unique to only one fine-tuning language and can thus be attributed to language bias. 
Yet, the values that shift are not consistent across test languages. This again shows that the pattern with which values shift, heavily vary based on the language used for probing. However, these results also indicate that, not only are certain languages more prone to change cultural perspective, there are per language also a specific set of values that are more prone to shift.

\paragraph{Are certain WVS categories more prone to domain bias?}
In Section~\ref{sec:prob_results}, we found that the pretrained models of different sizes agree more on certain WVS categories. Thus, we test whether the impact of the domain source will be more visible when studying results per category. In Figure~\ref{fig:radar_cats}, we report the Pearson correlation between the percentages of unchanged values per WVS category for each data source pair averaged across fine-tuning languages. We find that overall Tanzil and Bible tend to score higher compared to Flores-101. Yet, the lowest correlations between data sources are reported for religious values. This suggests that the different religious biases of PBC and Tanzil do in fact have a different effect on the value shifts.

\begin{figure*}[!t]
     \scalebox{0.9}{
      \includegraphics[width=0.49\linewidth, height=5.2cm]{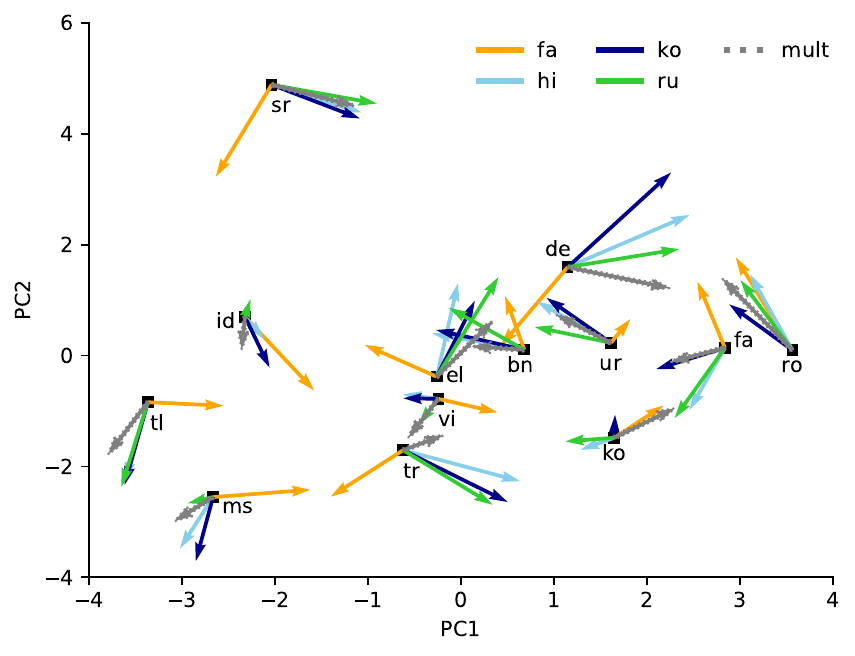}} 
      \scalebox{0.9}{
     \includegraphics[width=0.49\linewidth, height=5.2cm]{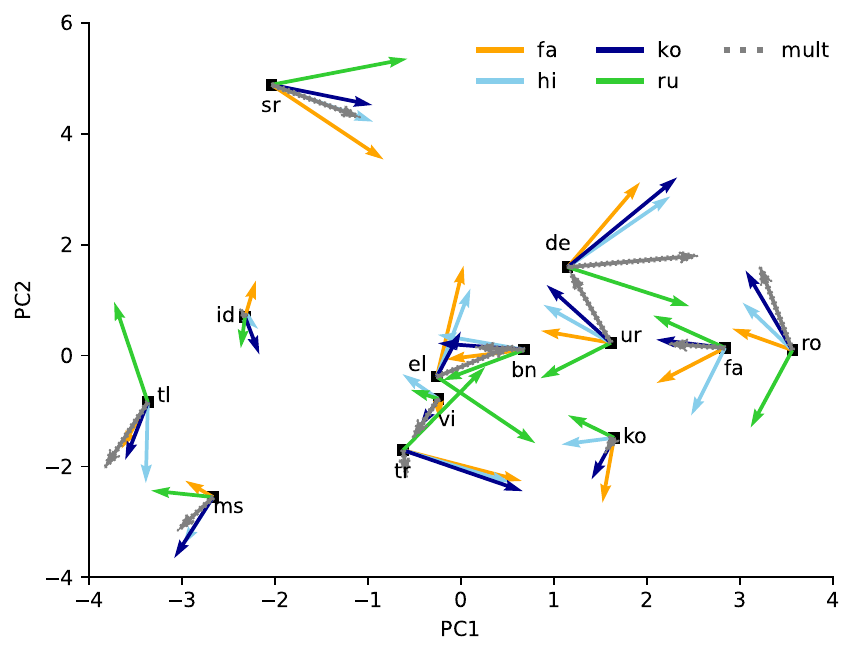}}
     \vspace{-0.1cm}
    \caption{Starting from the cultural profiles extracted from pretrained mT5-small, the image depicts in which direction each test language changes depending on the language selected for fine-tuning on PBC (left) and Tanzil (right). The cultural profiles are projected down to 2-dimensions using PCA~\citep{bro2014principal}. }
    \label{fig:change_dir}
\end{figure*}

\paragraph{Do we just need more fine-tuning examples?}
 A natural follow up question is whether language or domain bias becomes more prevalent when using a larger corpus during fine-tuning. We find that increasing our training size from 2K to 10K samples does tend to further increase the amount of value shifts (yet, it can also decrease, e.g. for German when fine-tuning in Farsi or Russian). Importantly, however, this does not considerably change the patterns between PBC and Tanzil. Moreover, note that a substantial amount of values shift after fine-tuning on 2K samples, which shows that the fine-tuning procedure has made an impact. 
\subsubsection{Does the direction of cultural change differ depending on the FT setup?}\label{sec:whatdir}
As we saw that the amount of changes are similar across fine-tuning languages and sources, we now instead study the effect that each fine-tuning setup has on the cultural profiles. In Figure~\ref{fig:change_dir}, we plot in which direction the cultural profiles of the pretrained model changed depending on the fine-tuning language used. In accordance with our previous results, we find that the direction of change is mostly dependent on the language used for testing, as most fine-tuning languages point in similar directions. However, we do see some differences when comparing results across datasets. For instance, we see that for Indonesian and Korean, the fine-tuning languages seem to steer the cultural bias into different directions depending on the dataset. Moreover, for PBC we find that across many test languages, fine-tuning in Farsi steers the model in a different direction compared to the other languages. Thus, while the amount of value shifts are only weakly affected by fine-tuning language and source, these results indicate that they can have a strong effect on the overall cultural bias of the model.

\subsubsection{Does multilingual FT affect cultural values differently?}\label{sec:monomulti}
In the previous sections, we studied the effect of monolingual fine-tuning. However, multilingual models are jointly trained on multiple languages, which further complicates which values the model should pick up on. Thus, we now test to what extent cross-language influence during multilingual fine-tuning affects the cultural bias of the models differently compared to monolingual fine-tuning. In Table~\ref{tab:monomult}, we report the amount of unchanged values after multilingual and monolingual fine-tuning. We find that the effect of multilingual fine-tuning is for many languages approximately similar to the average scores obtained across fine-tuning languages in a monolingual set-up (this pattern holds for models of varying sizes, see Appendix~\ref{app:robustness}). In addition, in Figure~\ref{fig:change_dir} we see that the direction in which the cultural profiles change, does not deviate much from monolingual fine-tuning in most cases. We suspect that the results for multilingual fine-tuning are similar because the fine-tuning languages in any case tend to behave similarly. Thus, when using them interchangeably it has a limited further effect on the predictions. 

\begin{table}[t!]
    \centering   
    \setlength{\tabcolsep}{4pt}
    \scalebox{0.65}{
    \begin{tabular}{r | c c c c c c c c c c c c c }
    \toprule
                 & bn & de & el & fa & id & ko &ms & ro& sr & tl & tr & ur & vi \\
            \midrule 
            mult & 76&74 & 74& 72 & 64& 67& 72& 74&65 & 69& 71& 41& 78 \\
            mono-avg  & 81&75 & 74&  72&  69& 71& 69 & 82& 73& 74& 73&  48 & 70\\
\bottomrule
\end{tabular}}
    \caption{Percentage of unchanged values after multilingual and monolingual fine-tuning.}
    \label{tab:monomult}
\end{table}

\section{Correlation with human survey results}
In Section~\ref{sec:prob_results}, we studied which cultural values were encoded in the pretrained LM, and in Section~\ref{res:shifts}, how much and in which direction these could change after fine-tuning. We now test whether the changes we observed led the model to be steered into a direction that is better aligned to real human values. Thus, 
we compute how much the Spearman correlation between the ground truth profiles and the pretrained LM changed after fine-tuning. To select culturally diverse countries to test alignment to, we first compute cosine similarities between the ground truth profiles of our 13 test languages, and found that the profiles from Germany and Pakistan (0.84 similarity) and Vietnam and Serbia (0.88 similarity) deviated the most.


\paragraph{How does the alignment between test languages and human values change?}
In Table~\ref{tab:human-corr}, we report the change in correlations averaged over fine-tuning languages. In Figure~\ref{fig:change_dir} we saw that, depending on the test language, fine-tuning changed the cultural information in different directions. We now see that this mostly leads to a better alignment to human data, regardless of domain source. For instance, fine-tuning on Tanzil and PBC on average increases correlation with all countries' data for Tagalog. This suggests that the model is overall pushed closer to human values. Moreover, test languages whose profiles pointed in different directions across datasets (e.g., Korean, Indonesian and Serbian), are now also affected differently in their alignment to human values. Yet, when looking at the absolute values for PBC and Tanzil, the correlations remain low across the board, showing that the model is still poorly aligned to human values. 

\begin{table}[!t]
    \centering   
    \setlength{\tabcolsep}{2pt}
    \scalebox{0.6}{
    \begin{tabular}{r  ccccccccccccc }
    \toprule
                 & bn &  de & ur & el & id & ko & ms & ro& sr & tl& tr& vi& fa  \\
        \midrule        
           & \multicolumn{13}{c}{PBC} \\
           \hline
          DE     & 
        \cellcolor{greenish}+.02 &\cellcolor{greenish}  +.11 &\cellcolor{greenish} +.03 &\cellcolor{greenish} +.04 & \cellcolor{greenish}+.09 & \cellcolor{orangy}-.03 &\cellcolor{greenish} +.12 &\cellcolor{greenish} +.01&\cellcolor{orangy} -.05 &\cellcolor{greenish} +.12& \cellcolor{greenish}+.04&\cellcolor{greenish} +.15& \cellcolor{orangy}-.03  \\
          PK    &  \cellcolor{orangy}-.13&\cellcolor{greenish}  +.02 &\cellcolor{orangy} -.11 & \cellcolor{greenish}+.16 &\cellcolor{greenish} +.14 & \cellcolor{greenish}+.13 &\cellcolor{greenish} +.18&\cellcolor{orangy} -.08 &\cellcolor{greenish} +.01&\cellcolor{greenish} +.23&\cellcolor{greenish} +.15&\cellcolor{orangy} -.02&\cellcolor{orangy} -.14  \\
          SR   & \cellcolor{orangy}-.06 & \cellcolor{greenish} +.26 &\cellcolor{orangy} -.06 &\cellcolor{greenish} +.08 & \cellcolor{greenish}+.04 &\cellcolor{orangy} -.01 &\cellcolor{greenish} +.01 & 0 &\cellcolor{orangy} -.08 & \cellcolor{greenish}+.16& \cellcolor{greenish}+.09& 0 &\cellcolor{orangy} -.07  \\
          VI    &\cellcolor{orangy} -.08 & \cellcolor{orangy} -.18 & \cellcolor{orangy}-.11 & 0 & \cellcolor{greenish}+.01 &\cellcolor{orangy} -.01 &\cellcolor{orangy} -.03 &\cellcolor{orangy} -.09& \cellcolor{greenish}+.01 & \cellcolor{greenish}+.12&  0 & \cellcolor{orangy}-.14&\cellcolor{greenish} +.04  \\
          \hline
           & \multicolumn{13}{c}{Tanzil} \\
           \hline
          DE     & 
        \cellcolor{orangy}-.02 &   \cellcolor{greenish}+.18 &  \cellcolor{greenish}+.06 &  \cellcolor{greenish}+.02 &  \cellcolor{greenish}+.10 & \cellcolor{orangy}-.04 & \cellcolor{greenish} +.12 &\cellcolor{greenish} +.08&\cellcolor{orangy} -.06 &\cellcolor{greenish} +.14& \cellcolor{orangy}-.02&  \cellcolor{greenish}+.19&\cellcolor{orangy} -.02  \\
          PK    &\cellcolor{orangy} -.16&  \cellcolor{greenish} +.02 &\cellcolor{orangy} -.08 & \cellcolor{greenish} +.16 & \cellcolor{greenish} +.12 & \cellcolor{greenish} +.13 & \cellcolor{greenish} +.11 &\cellcolor{orangy} -.08& 0 & \cellcolor{greenish} +.26& \cellcolor{greenish} +.18&\cellcolor{orangy} -.01&\cellcolor{orangy} -.14  \\
          SR   & \cellcolor{orangy}-.08 &  \cellcolor{greenish} +.03 &\cellcolor{orangy} -.06 & \cellcolor{greenish} +.09 & \cellcolor{greenish} +.03 &\cellcolor{orangy} -.04 &\cellcolor{orangy} -.01 & 0 &\cellcolor{orangy} -.06 &  \cellcolor{greenish}+.15& \cellcolor{greenish} +.05& \cellcolor{greenish} +.04&\cellcolor{orangy} -.06  \\
          VI    & \cellcolor{orangy}-.09 & \cellcolor{orangy} -.24 &\cellcolor{orangy} -.07 &  \cellcolor{greenish}+.01 & \cellcolor{orangy}-.08 &  \cellcolor{greenish}+.05 &\cellcolor{orangy} -.08 & \cellcolor{orangy}-.11& \cellcolor{greenish} +.05& \cellcolor{greenish} +.11& \cellcolor{greenish} +.04&\cellcolor{orangy} -.15 & \cellcolor{greenish} +.05  \\
        
\bottomrule
\end{tabular}}
    \caption{Change in alignment to the ground truth profiles for each country (DE, PK, SR, VI), measured by the difference in Spearman correlation. Results are averaged over fine-tuning languages.}
    \label{tab:human-corr}
\end{table}

\paragraph{How does fine-tuning affect the cultural similarity between test languages within a model?}
\begin{figure}[!t]
    \centering
    \scalebox{0.6}{
    \includegraphics[width=\linewidth]{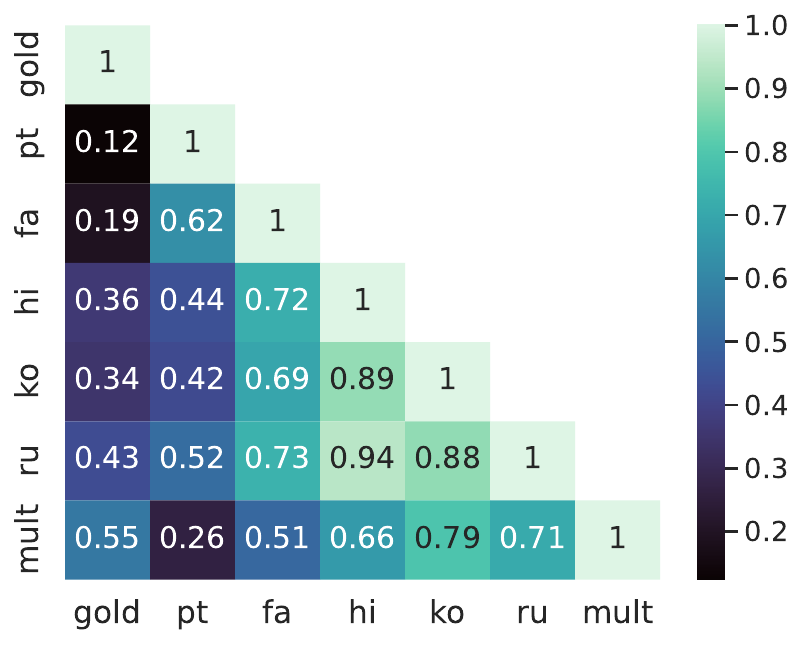}}
    \caption{Spearman correlation between the similarity matrices of the cultural profiles computed from the ground truth data, and pretrained and fine-tuned models.}
    \label{fig:rsa}
\end{figure}

Given that our models are poorly aligned to human data, we test whether at least the cultural similarities between different test languages 
correlate with those between real countries. For instance, do we find that the cultural profiles from Romanian and Serbian are more similar than those from Serbian and Urdu? To test this, we, for each model, compute a dissimilarity matrix between the cultural profiles of all language pairs using cosine similarity, and then use Spearman correlation to test how similar these matrices are across models~\citep{abnar2019blackbox}. In Figure~\ref{fig:rsa}, we find that while cultural relationships between languages in the pretrained LM are weakly correlated with human data, the alignment mostly increases after fine-tuning on PBC, (except for Farsi). Interestingly, multilingual fine-tuning results in the highest correlation with human data. The same result was found for Tanzil, see Appendix~\ref{app:changes}. When looking at the dissimilarity matrices for these models, we also find that the cultural profiles are more distinct, resulting in less similarity between language pairs. We suspect that, as a result of seeing multiple languages during training, various language-specific biases can be preserved and transmitted. In contrast, after monolingual fine-tuning, all languages are biased in one direction, resulting in very similar cultural profiles that do not preserve cross-cultural differences.


\section{Tracing cultural value shifts}

We found that fine-tuning languages have similar effects across test languages. Thus, as a complementary study, we test which training examples, and the languages they come from, influenced value shifts the most.  We use 
TRAK, a TDA method proposed by \citet{park2023trak}. 
We follow \citet{park2023trak} in treating the MLM objective as a multi-class classification problem, i.e.,\ framing it as a sequence of $v$-way classification problems over masked tokens, where $v$ is the vocabulary size. We use the TRAK library\footnote{\url{https://github.com/MadryLab/trak}}~for our implementation and project gradients down to 4096 dimensions, all other parameters are kept at default. See \citet{park2023trak} for a detailed explanation. Per value shift we analyze the top 100 most influential training examples.





\paragraph{Are value shifts influenced by the same training examples across fine-tuning and test languages?}
In Section~\ref{sec:prob_results}, we saw that value shifts were mostly not unique to one fine-tuning language. Thus, we test whether these shifts were actually influenced by the same training examples across fine-tuning languages. Interestingly, we find that from the most influential examples in each fine-tuning language that instigate the same value shift, only \textless $ 5\%$ are parallel sentences. Yet, when looking at the values that shift across test languages given the same fine-tuning language, we observe more consistency. 
For PBC, we find that across all language pairs, when fine-tuning in Farsi and Hindi, up to 20$\%$ of training examples are consistently relied upon across test languages, and for Korean and Russian 49 and 62$\%$ resp. These results suggest that the semantic content of fine-tuning data might not be the main reason behind the shifts. Instead, the model tends to rely more on the same training examples within a fine-tuning language irrespective of test language. 

\begin{figure}[!t]
    \centering
    \scalebox{1}{
    \includegraphics[width=\linewidth, height=2.5cm]{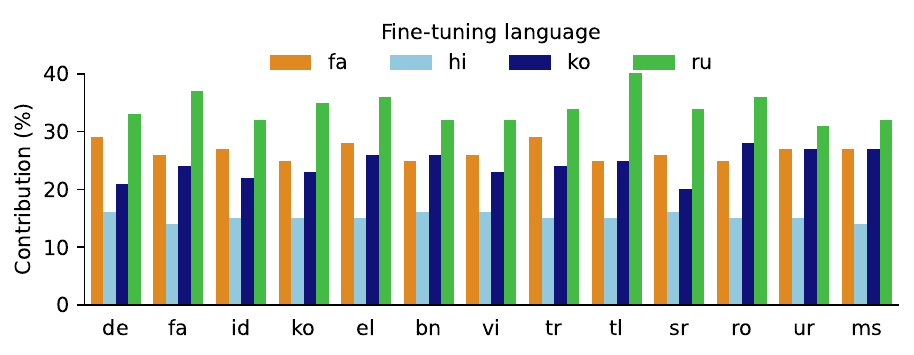}}
    \scalebox{1}{
     \includegraphics[width=\linewidth, height=2.5cm]{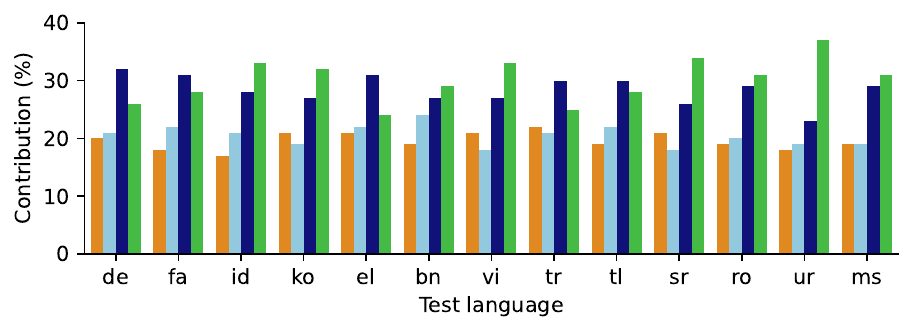}}
    \caption{The average percentage of training samples from each fine-tuning language that contributed to the top 100 training samples for a test language after multilingual fine-tuning on PBC (top) and Tanzil (bottom).}
    \label{fig:mult_trak}
\end{figure}
\paragraph{Which languages instigate the value shifts during multilingual fine-tuning?}
We use the approach from \citet{choenni2023languages} to quantify cross-language influence by the average percentage of training examples that each fine-tuning language contributes to the most influential examples for each test language. In Figure~\ref{fig:mult_trak}, we see that for PBC, Russian and Farsi have on average the largest influence across test languages. Interestingly, for Tanzil, we instead see that Russian and Korean contribute the most. Given that different trends across datasets could also be an artifact of randomness during fine-tuning, we repeat the experiment for PBC on a model fine-tuned with a different random seed, but confirm that the trends hold. While the large influence of Russian could be explained by the fact that it has the second largest dataset for pretraining, the results indicate that the influence of languages on value shifts during multilingual fine-tuning is dependent on the content of the fine-tuning data.

\section{Discussion and conclusion}

We studied to what extent fine-tuning languages and domain sources exert influence on cultural values encoded for a set of test languages in MLMs. In particular, we tested how different fine-tuning setups can change the overall cultural biases across test languages differently, and in which cases this leads the model to be better aligned to human values. We found that fine-tuning language and domain source play a minor, but visible, role in the amount of value shifts compared to size of the fine-tuning dataset. Moreover, results vary considerably across test languages. Still, different fine-tuning languages can cause cultural profiles of test languages to be steered into different directions, which leads to varying effects on the models' alignment to human values. In addition, we find that multilingual fine-tuning better preserves the human cultural similarities between test languages within a LM. 

Finally, our TDA analysis shows that while different fine-tuning languages can lead to the same value shifts, the training examples that are relied upon vary. This suggests that the semantic content of fine-tuning data might no be the main reason for the shifts. Instead, the model tends to rely on the same training examples within a fine-tuning language, and these examples have different effects on the manifestation of cultural values across test languages. Hence, future work on value alignment likely requires a different adaptation approach for each test language. While multilingual NLP has made big strides in the past years, the field of cross-cultural NLP is still in its infancy as many questions remain to be explored. We hope that our insights will inform future work on value 
alignment to enable more culturally-aware language technology.

\section*{Limitations}

While language and culture are closely connected \citep{kramsch2014language,  hovy2021importance}, we can not use these notions interchangeably \citep{hershcovich2022challenges}. For instance, even within a language many subcultures typically exist, and the idea that for instance “English”
carries a single set of values has been discarded \citep{paul2009cross}. At the same time, multiple languages can also carry a relatively homogeneous culture \citep{sahlgren2021s}.
While the languages were selected based on the criteria that its speakers can be primarily localized to a specific geographical region (and thus likely maintain their own cultural profile), we can not guarantee that all online texts in that language transmit the same cultural values.

Moreover, we were restricted in the choice of domain source and fine-tuning language combination due to a lack of available datasets that contain a sufficient amount of multi-parallel data for fine-tuning. While we could, for instance, use many languages from the Flores-101 dataset, each language only contains approximately 2K multi-parallel sentences. While PBC and Tanzil contain different religious biases, it could also be argued that these data sources are in fact not substantially dissimilar. 

Finally, while we use data from one of the most popular cross-cultural value questionnaires from social science, i.e. WVS, it also has its shortcomings. In particular, similar to how languages do not contain a single culture, it is also questionable to map an entire country to a single set of cultural values. This is particularly true for countries with many immigrants of different cultural backgrounds. As such, there are also different subcultures within a country, making it not obvious that we should simply map a MLM to a countries' cultural values based on the WVS data. This also further complicates how we should interpret an alignment between language and country as it can easily be mismatched. Thus, in future work, researchers from various disciplines should investigate and discuss what an ideal cultural alignment for a MLM should look like in practice. 

\section*{Ethical considerations}
All data sources used in this study are publicly available. While we acknowledge that automatically analyzing religious texts can be a sensitive topic, we do not draw any conclusions based on the content of those data sources in this work nor do we provide examples from the texts directly. 
Moreover, while we test for cultural alignment to human data in this study, we recognize that languages can not simply be mapped to single countries and therefore it is not always straightforward to decide which human values the model should align to in practice. As such, we leave this question in the middle, and rather just explore to what extent we can influence the cultural profiles of MLMs. 
\section*{Acknowledgements}
Rochelle Choenni is supported by a Google PhD Fellowship. The work of Anne Lauscher is funded under the Excellence Strategy of the German Federal Government and the Federal States.


\appendix

\onecolumn

\section{Agreement between pretrained LMs}\label{app:pretrained}
\begin{figure}[H]
    \centering
\includegraphics[width=\linewidth]{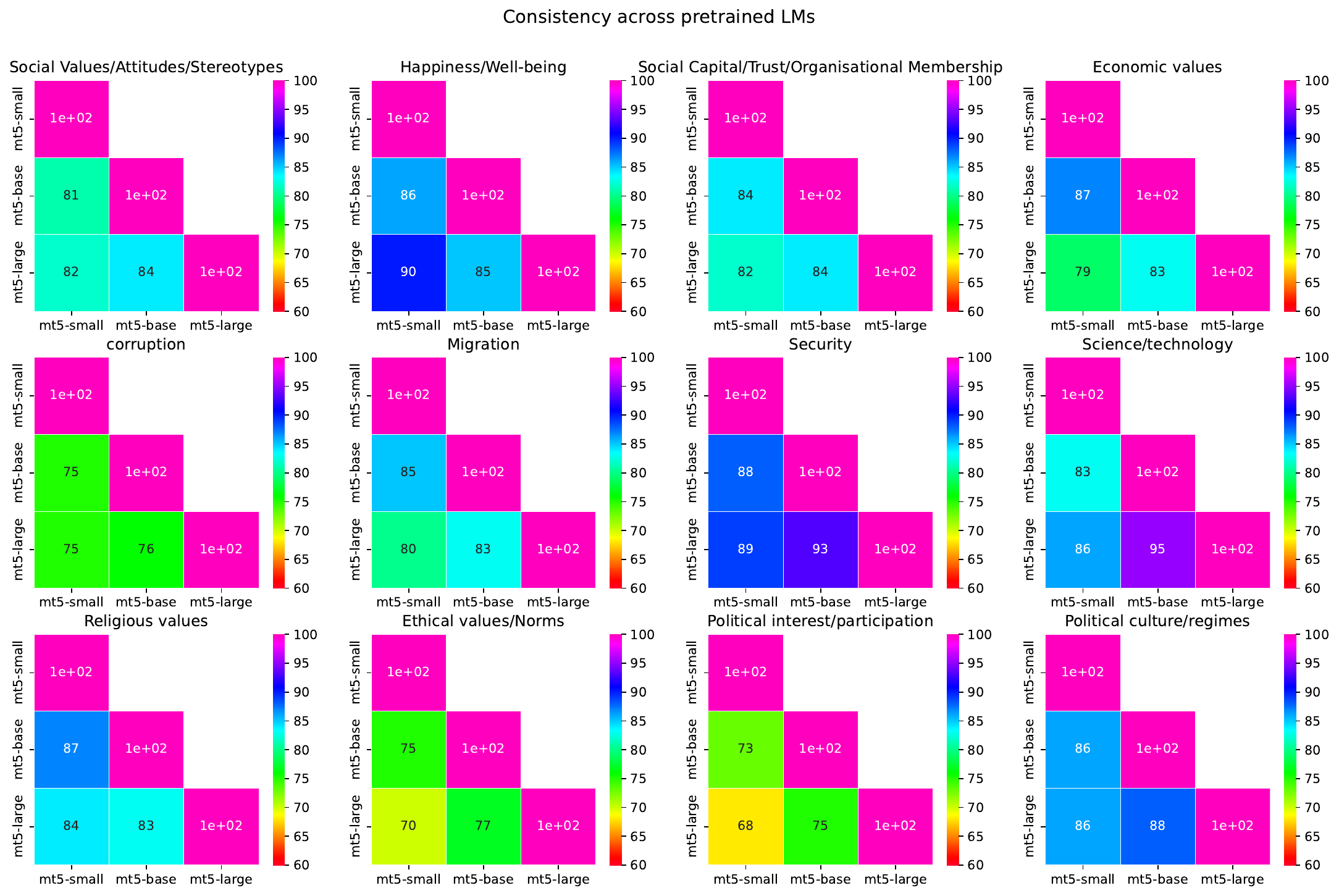}

    \includegraphics[width=\linewidth]{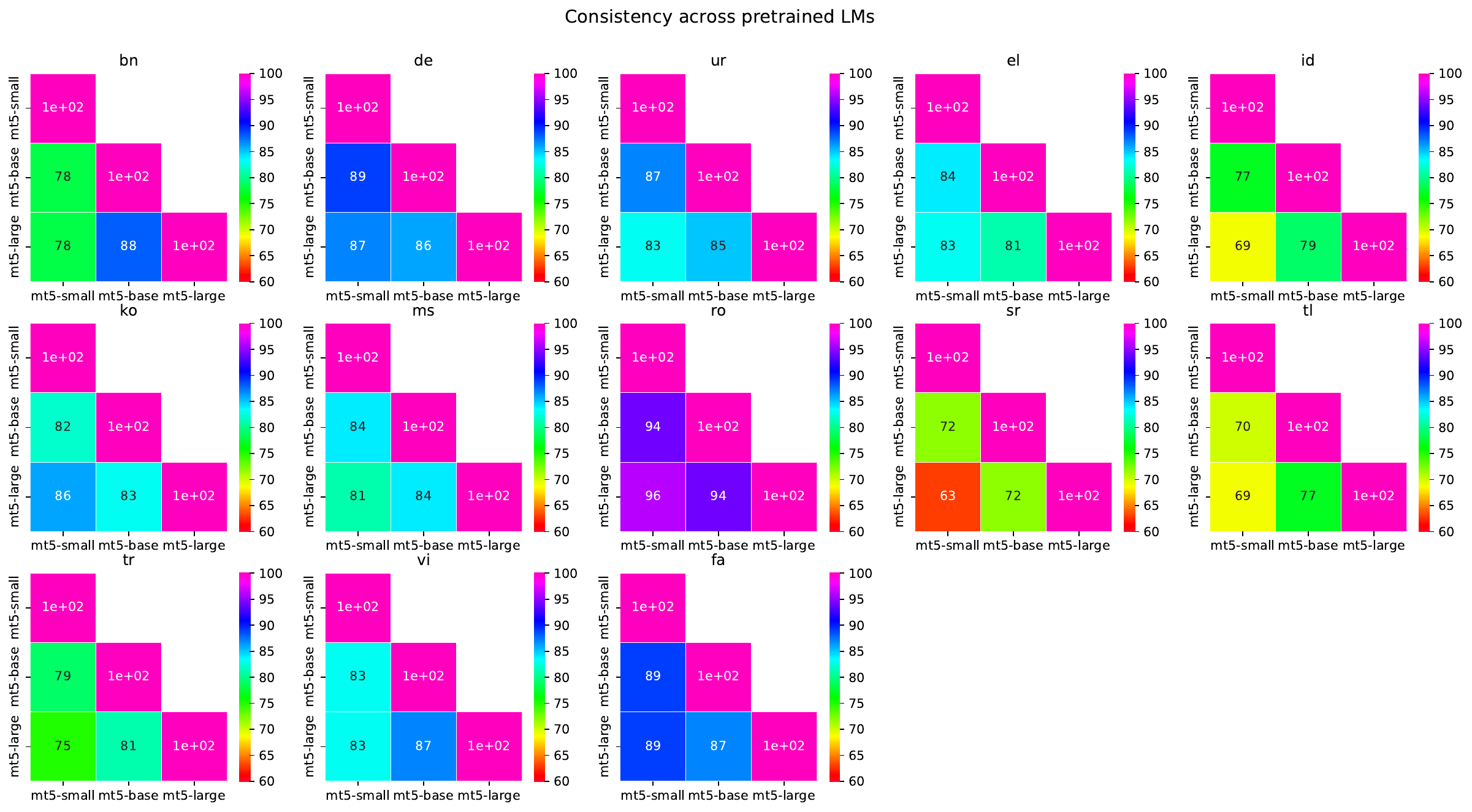}
    \caption{The percentage of survey questions for which pretrained mT5-small models with different number of parameters were in agreement about the answer they outputted. We show the percentage per category averaged over test languages (top) and the percentage per test language averaged over categories (bottom)}
    \label{fig:pretrained_agre}
\end{figure}
\clearpage

\section{Fine-tuning details}\label{app:FT}
We use a 80/20 train/development split, a learning rate of 5e-5, the AdamW optimizer and a batch size of 8, and train for 5 epochs. We query the models through the Huggingface Library and use its Trainer class with default hyperparameters for fine-tuning \citep{wolf2019huggingface}. Moereover, all fine-tuning and tracing experiments are ran on a NVIDIA A100-SXM4 GPU with 40GB memory.

\section{Percentage of unchanged value predictions after fine-tuning }\label{app:robustness}
\vspace{-1cm}
\begin{figure}[H]
    \centering
    \includegraphics[width=\linewidth]{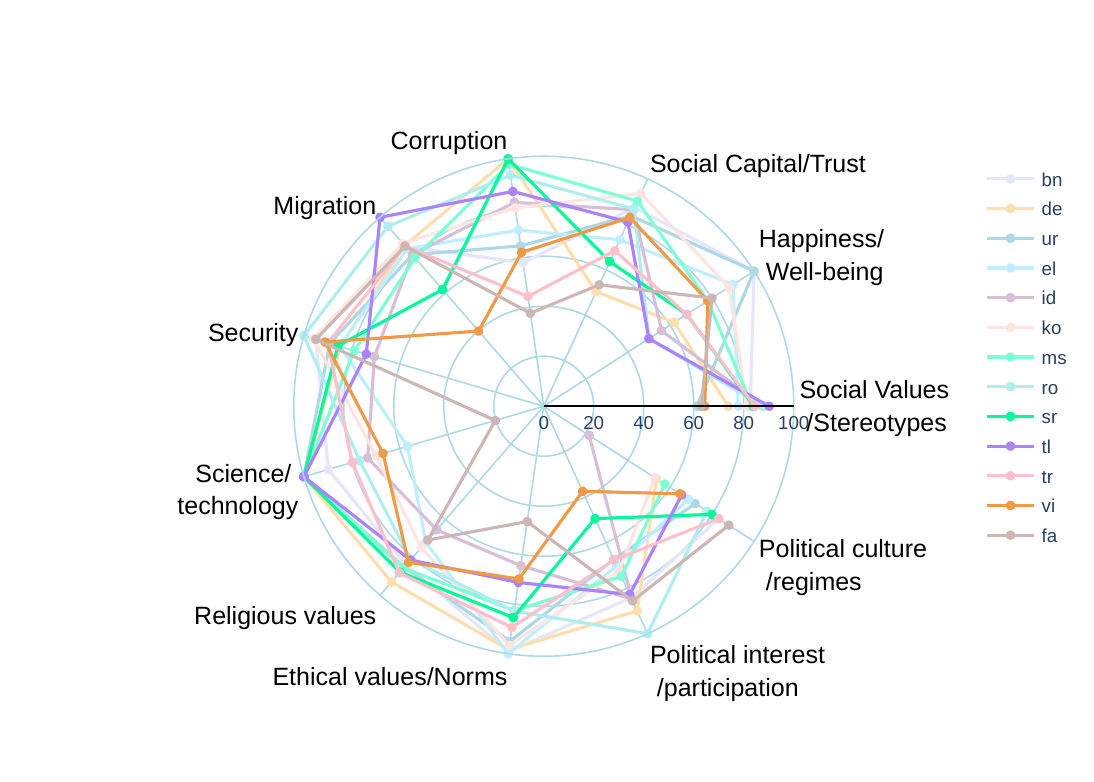}
    \caption{The percentage of unchanged values per test language for each WVS category after fine-tuning on PBC. Results are averaged across fine-tuning languages.}
    \label{fig:radar_langs}
\end{figure}

\begin{figure}[H]
    \centering
    \includegraphics[width=0.33\linewidth]{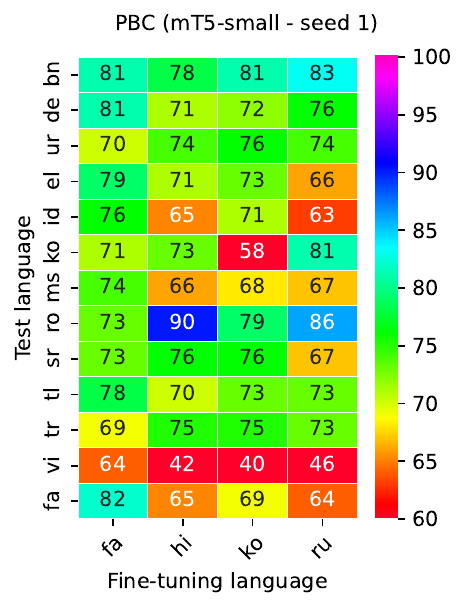}
        \includegraphics[width=0.33\linewidth]{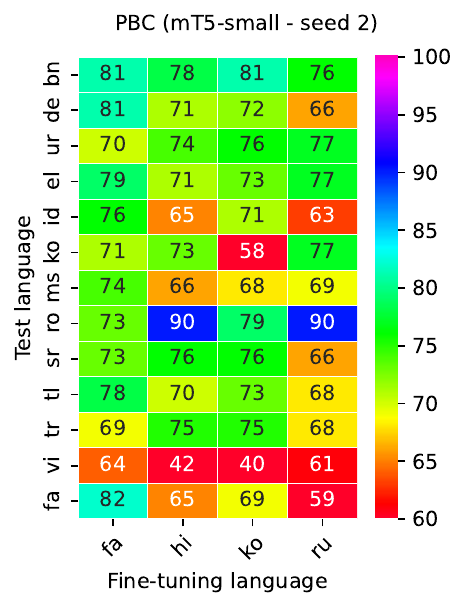}
         \includegraphics[width=0.33\linewidth]{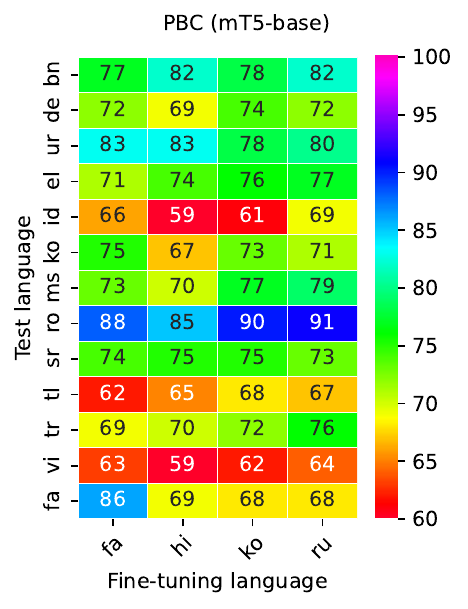}
    \includegraphics[width=0.33\linewidth]{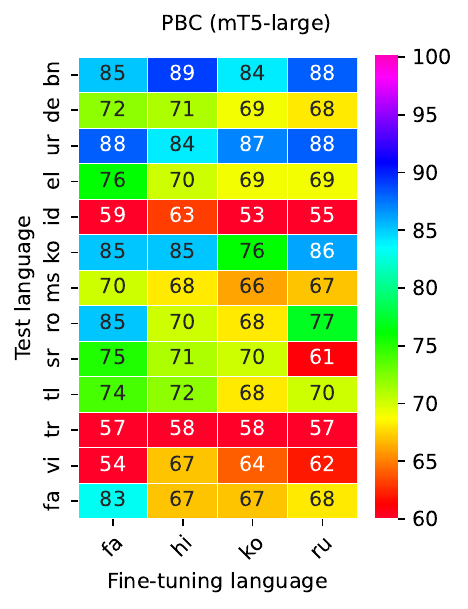}
    \caption{The percentage of of unchanged values after fine-tuning mT5-small on 10K sentences from PBC. We see the effect of using 2 different random seeds during fine-tuning, and the effect of using different model sizes i.e., mT5-base and mT5-large.}
    \label{fig:pretrained_const}
\end{figure}

\paragraph{Robustness analysis}\label{sec:robust}
We separately fine-tune the mT5-small in each language on PBC with 3 random seeds, and show results of two seeds in Figure~\ref{fig:pretrained_const}. We find that both across different random seeds for the same model and across mT5 of different model sizes, the amount of predictions that change after fine-tuning compared to the pretrained LM are relatively similar. However, we do see that for mT5-large, language-wise patterns become more distinct. For instance, across all fine-tuning languages, we see that the predictions for Bengali and Urdu remain more robust compared to the smaller models. For Turkish and Indonesian we see an opposite effect where instead across all fine-tuning languages the predicted values tend to change more. Similarly, we compared performance to fine-tuning using only 1 training epoch, while this slightly reduces the amount of value shifts, the overall patterns did not change considerably.

\begin{figure}[H]
    \centering
    \scalebox{0.6}{
    \includegraphics[width=\linewidth]{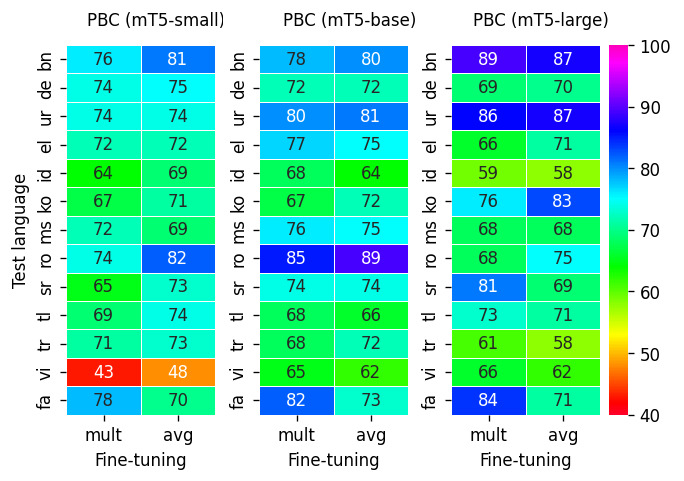}}
    \caption{The percentage of survey questions for which the value prediction did not change after multilingual fine-tuning and on average when monolingual fine-tuning using the same languages.}
    \label{fig:mult}
\end{figure}

\section{Changes to cultural profiles after fine-tuning}\label{app:changes}

\begin{figure}[H]
    \centering
    \includegraphics[width=0.49\linewidth]{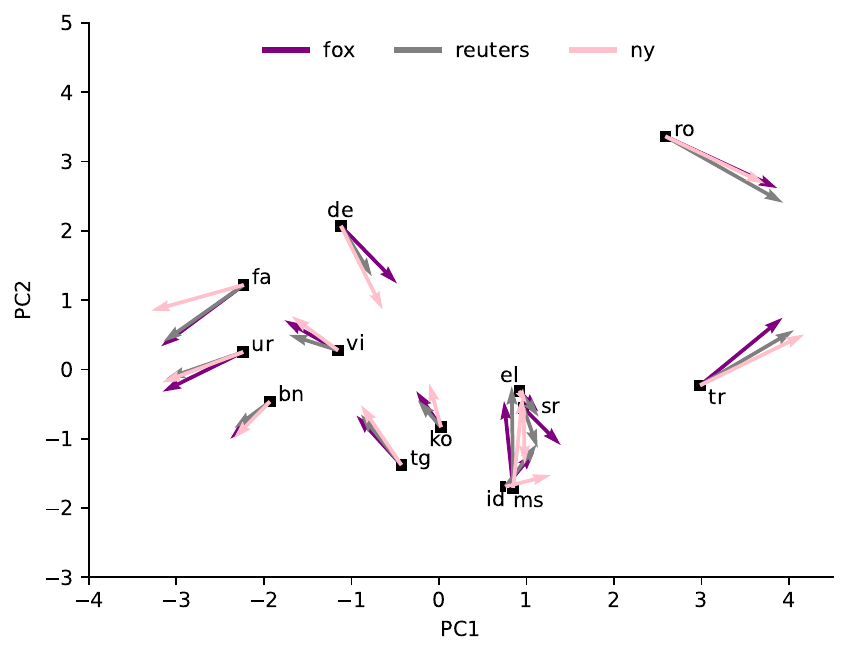}
        \includegraphics[width=0.49\linewidth]{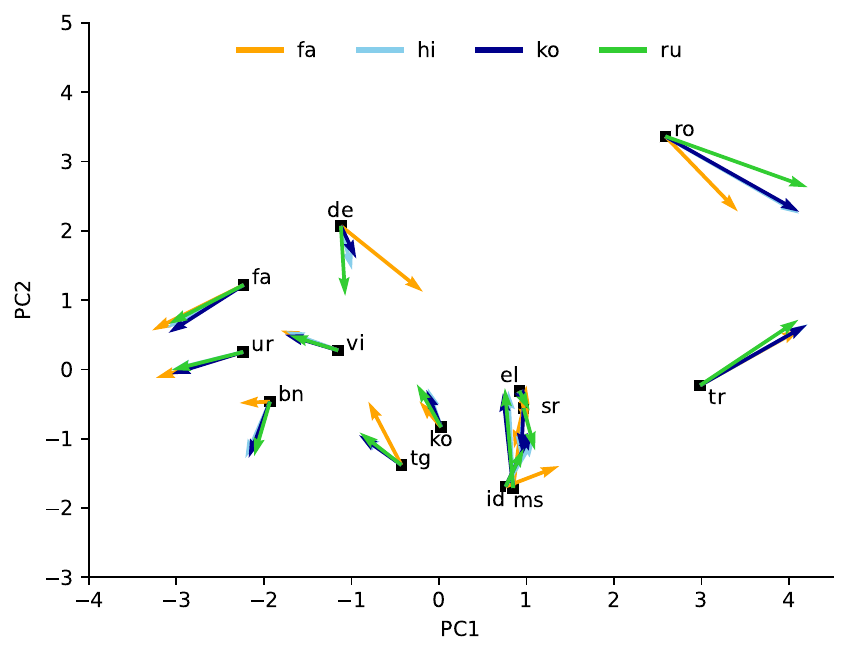}
    \includegraphics[width=0.49\linewidth]{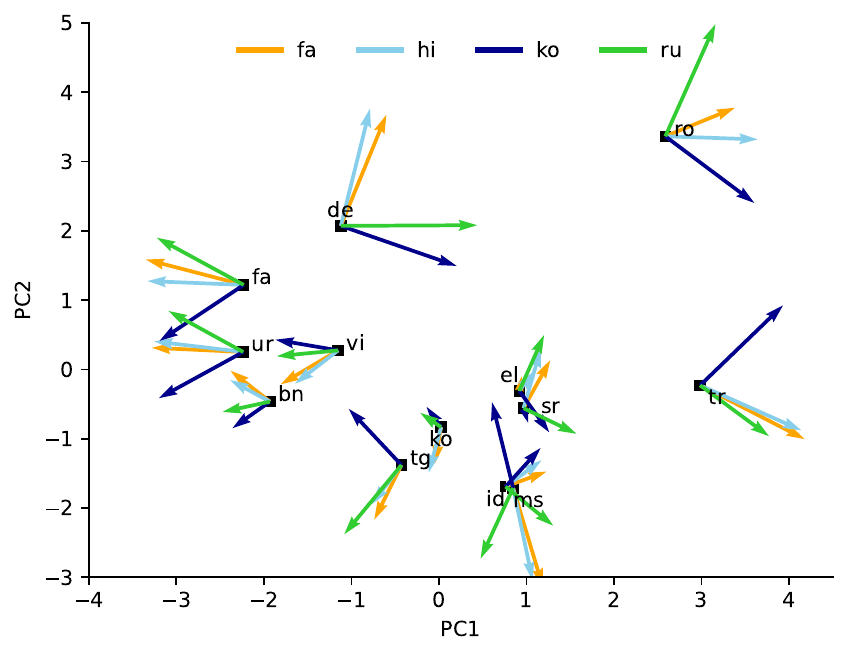}
    \includegraphics[width=0.49\linewidth]{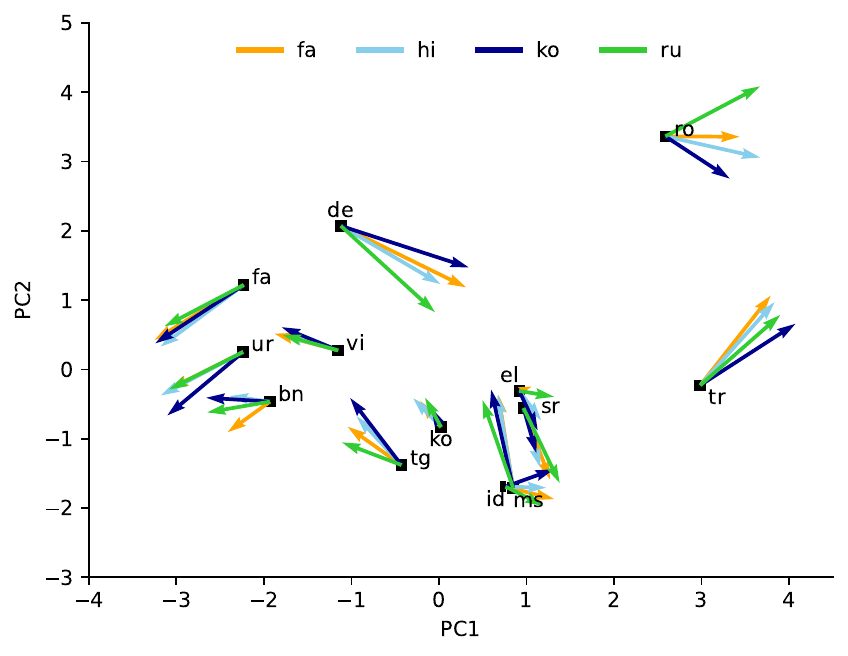}
    \caption{Starting from the cultural profiles extracted from pretrained mT5-small, the image depicts into which direction each test language changes depending on the source selected for fine-tuning on 2K sentences: news articles (top left), Flores (top right), PBC (bottom left) and Tanzil (bottom right. The cultural profiles are projected down to 2-dimensions using PCA.}
    \label{fig:enter-label}
\end{figure}

\begin{figure}[H]
    \centering
     \includegraphics[width=0.4\linewidth]{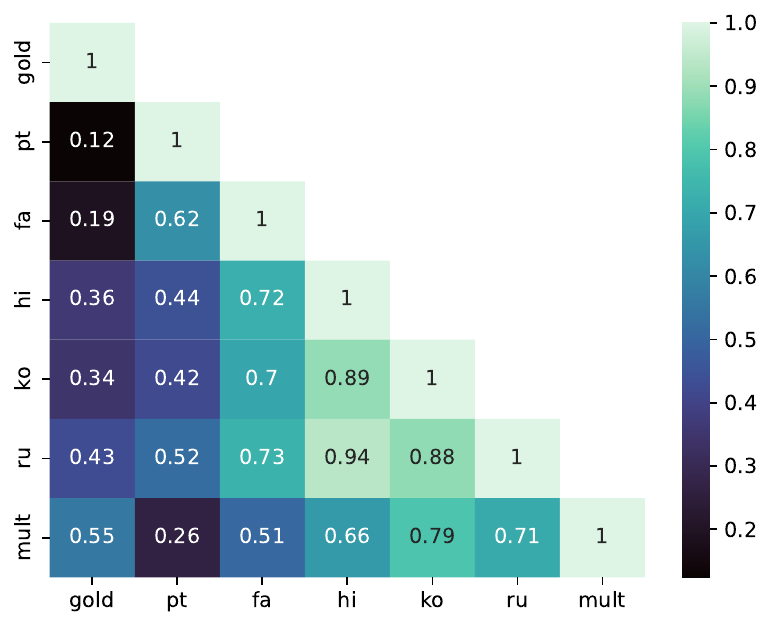}
    \includegraphics[width=0.4\linewidth]{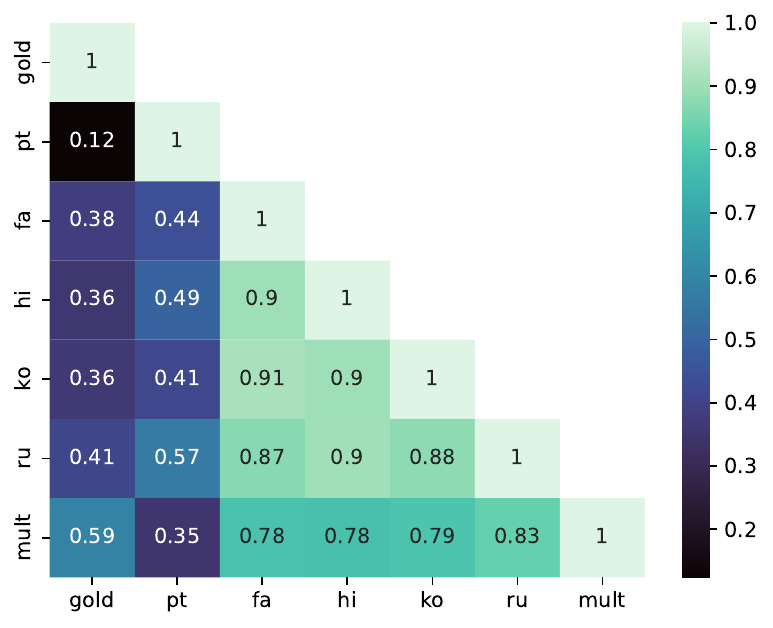}
    \caption{Spearman correlation between the similarity matrices of the cultural profiles computed from the ground truth data, and pretrained and the models fine-tuned on PBC (left) and Tanzil (right).}
    \label{fig:quran_rsa}
\end{figure}

\section{Ground truth cultural profiles}\label{app:WVS}

\begin{figure}[H]
    \centering
    \scalebox{1}{
    \includegraphics[width=0.49\linewidth]{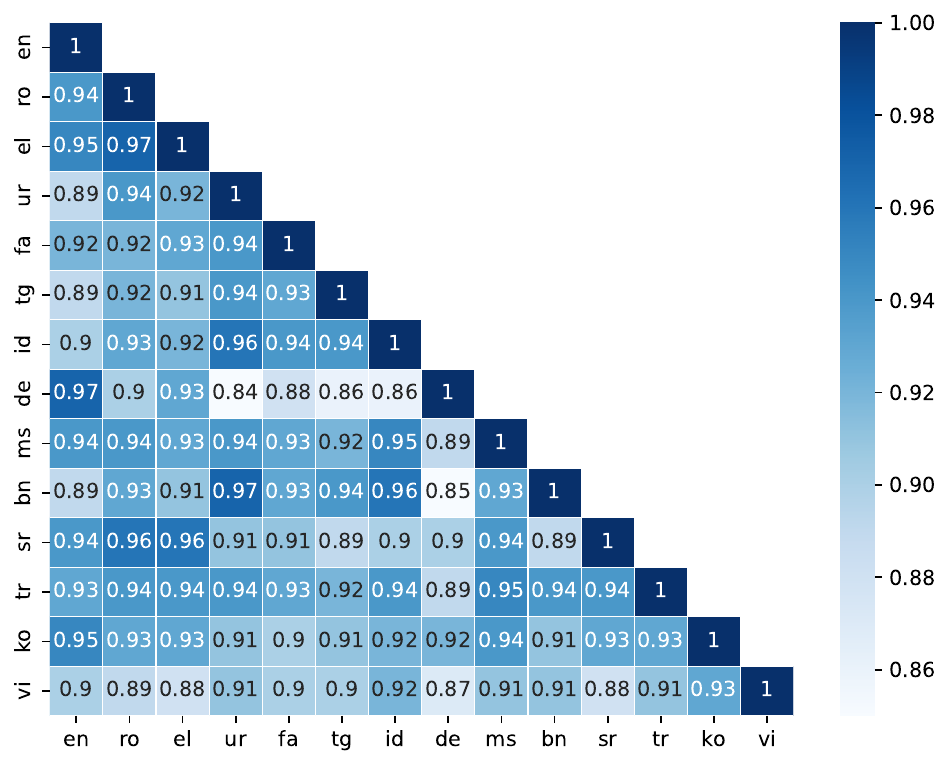}}
      \includegraphics[width=0.49\linewidth]{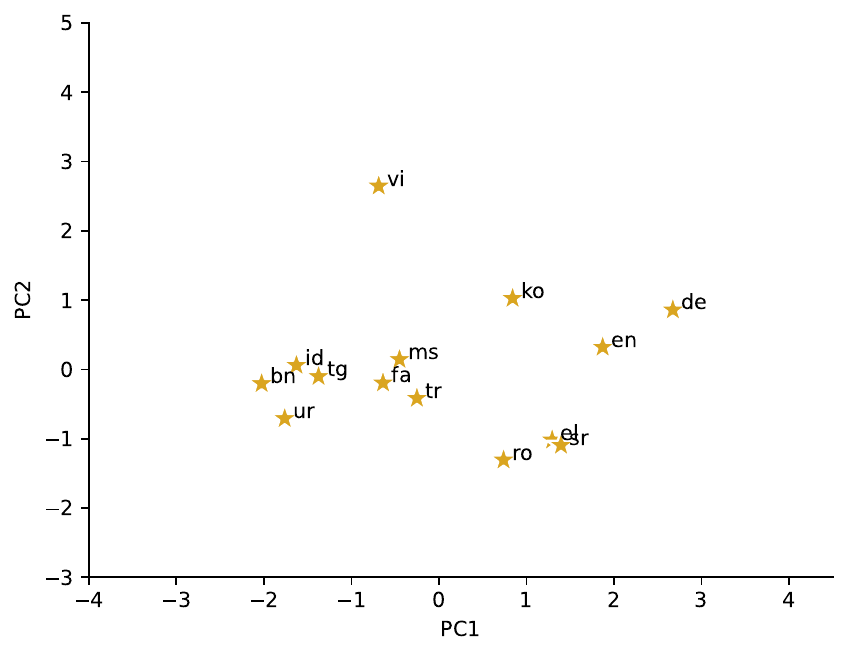}
    \caption{Left: The similarity between the cultural profiles of different countries according to the WVS survey results. Right: the ground truth profiles from each country projected down to 2 dimensions using PCA. }
    \label{fig:wvs_countries}
\end{figure}

\section{TRAK analysis}
\begin{figure}[H]
    \centering
    \includegraphics[width=0.49\linewidth]{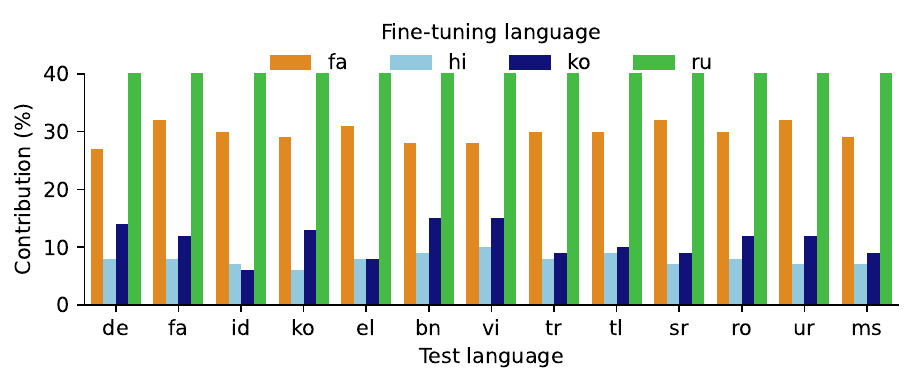}
     \includegraphics[width=0.49\linewidth]{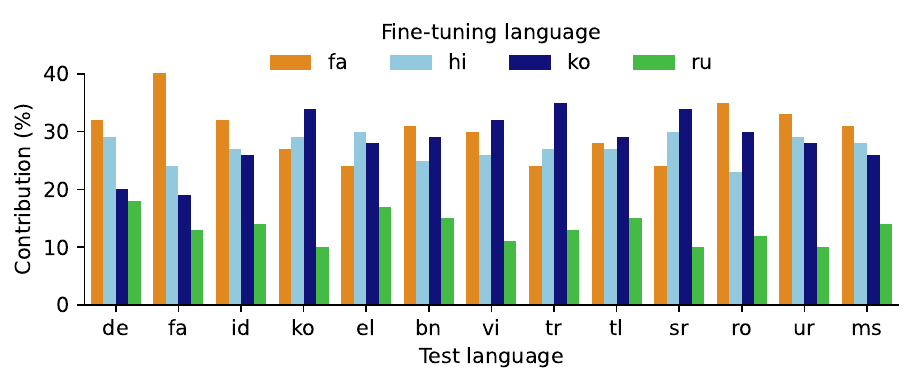}
    \caption{The average percentage of training samples from each fine-tuning language that contributed to the top 100 \emph{contradicting} training samples for a test language after multilingual fine-tuning on PBC (left) and Tanzil (right).}
    \label{fig:mult_trak_app}
\end{figure}

\end{document}